\documentclass{article}
\pdfoutput=1
 \PassOptionsToPackage{numbers, compress}{natbib}

\usepackage[preprint]{neurips_2021}

\usepackage[utf8]{inputenc} 
\usepackage[T1]{fontenc}    
\usepackage{hyperref}       
\usepackage{url}            
\usepackage{booktabs}       
\usepackage{amsfonts}       
\usepackage{nicefrac}       
\usepackage{microtype}      
\usepackage{xcolor}         

\usepackage[utf8]{inputenc}
\usepackage{amsmath}
\usepackage{amssymb}
\usepackage{graphicx}
\usepackage{cancel}
\usepackage{dsfont}
\usepackage{algorithm}
\usepackage{algpseudocode}
\usepackage{tikz}
\usepackage{bbm}
\usepackage{dsfont}
\usepackage{graphicx}
\usepackage{amsthm}
\usepackage{dsfont}
\usepackage{graphicx}
\usepackage{subcaption}
\usepackage{multirow}
\usepackage{placeins}

\usepackage{enumerate}
\usepackage{xspace}
\usepackage{romannum}
\usepackage{xcolor,colortbl}
\usepackage{tabularx}
\usepackage[labelfont=bf]{caption}

\usepackage{multicol}
\usepackage{wrapfig}

\DeclareMathOperator*{\argmax}{arg\,max}

\newcommand{\maxim}[2]{\ensuremath{\underset{#2}{#1}}}
\newcommand{\nrm}[2]{\ensuremath{{\parallel}{#2}{\parallel_{#1}}}}

\newcommand{\expt}[2]{\ensuremath{\underset{{#1}}{\mathbb E}{\{{#2}\}}}}

\newcommand{\entropy}[1]{\ensuremath{\mathcal{H}({#1})}\xspace}

\theoremstyle{plain}
\newtheorem{thm}{\protect\theoremname}
\theoremstyle{remark}

\theoremstyle{plain}

\theoremstyle{definition}
\newtheorem{defn}{\protect\definitionname}
\providecommand{\claimname}{Claim}
\providecommand{\definitionname}{Definition}
\providecommand{\lemmaname}{Lemma}
\providecommand{\theoremname}{Theorem}
\newtheorem{assumption}{Assumption}

\newtheorem{case}{Case}
\newcount\savedcase
\newenvironment{subcases}{%
	\savedcase=\value{case}%
	\edef\prevthecase{\thecase}%
	\setcounter{case}{0}%
	\renewcommand\thecase{\prevthecase.\arabic{case}}%
}
{%
	\setcounter{case}{\savedcase}%
}

\newcommand{\bydef}{\triangleq}
\newcommand{\probd}{\mathbb{P}}

\newcommand{\sep}{ | }

\newcommand{\lb}{\mathcal{LB}}
\newcommand{\ub}{\mathcal{UB}} 

\newcommand{\lucb}{\underline{\text{UCB}}}
\newcommand{\uucb}{\overline{\text{UCB}}}

\newcommand{\hist}{h_k = \{b_0, a_0, z_1, ...a_{k-1}, z_k\}}

\newcommand{\mot}[3]{\probd_T({#3} \sep {#2}, {#1})}
\newcommand{\observ}[2]{\probd_Z({#2} \sep {#1})}

\title{Simplified Belief-Dependent Reward MCTS Planning with Guaranteed Tree Consistency}

\author{Ori Sztyglic$^{*,1}$, Andrey Zhitnikov$^{*,2}$, Vadim Indelman$^3$ \\
	$^*$Equal contribution. \\
	$^1$Department of Computer Science \quad $^2$Technion Autonomous Systems Program \\
	$^3$Department of Aerospace Engineering\\
	Technion - Israel Institute of Technology, Haifa 32000, Israel\\
	\small{\texttt{ori.sztyglic@gmail.com, andreyz@campus.technion.ac.il, vadim.indelman@technion.ac.il}}}

\begin{document}
	
	\maketitle
	
	
	\begin{abstract}
Partially Observable Markov Decision Processes (POMDPs) are notoriously hard to solve. Most advanced state-of-the-art online solvers leverage ideas of Monte Carlo Tree Search (MCTS). These solvers rapidly converge to the most promising branches of the belief tree, avoiding the suboptimal sections. Most of these algorithms are designed to utilize straightforward access to the state reward and assume the belief-dependent reward is nothing but expectation over the state reward.  Thus, they are inapplicable to a more general and essential setting of belief-dependent rewards.  One example of such reward is differential entropy approximated using a set of weighted particles of the belief.  Such an information-theoretic reward introduces a significant computational burden.  In this paper, we embed the paradigm of simplification into the MCTS algorithm. In particular, we present \emph{Simplified Information-Theoretic Particle Filter Tree} (SITH-PFT), a novel variant to the MCTS algorithm that considers information-theoretic rewards but avoids the need to calculate them completely. We replace the costly calculation of information-theoretic rewards with adaptive upper and lower bounds.  These bounds are easy to calculate and tightened only by the demand of our algorithm. Crucially, we guarantee precisely the same belief tree and solution that would be obtained by MCTS, which explicitly calculates the original information-theoretic rewards. Our approach is general; namely, any converging to the reward bounds can be easily plugged-in to achieve substantial speedup without any loss in performance.
\end{abstract}

	
\section{Introduction}
\subsection{POMDPs}
\begin{figure}[t]
	\centering
	\includegraphics[width=\textwidth]{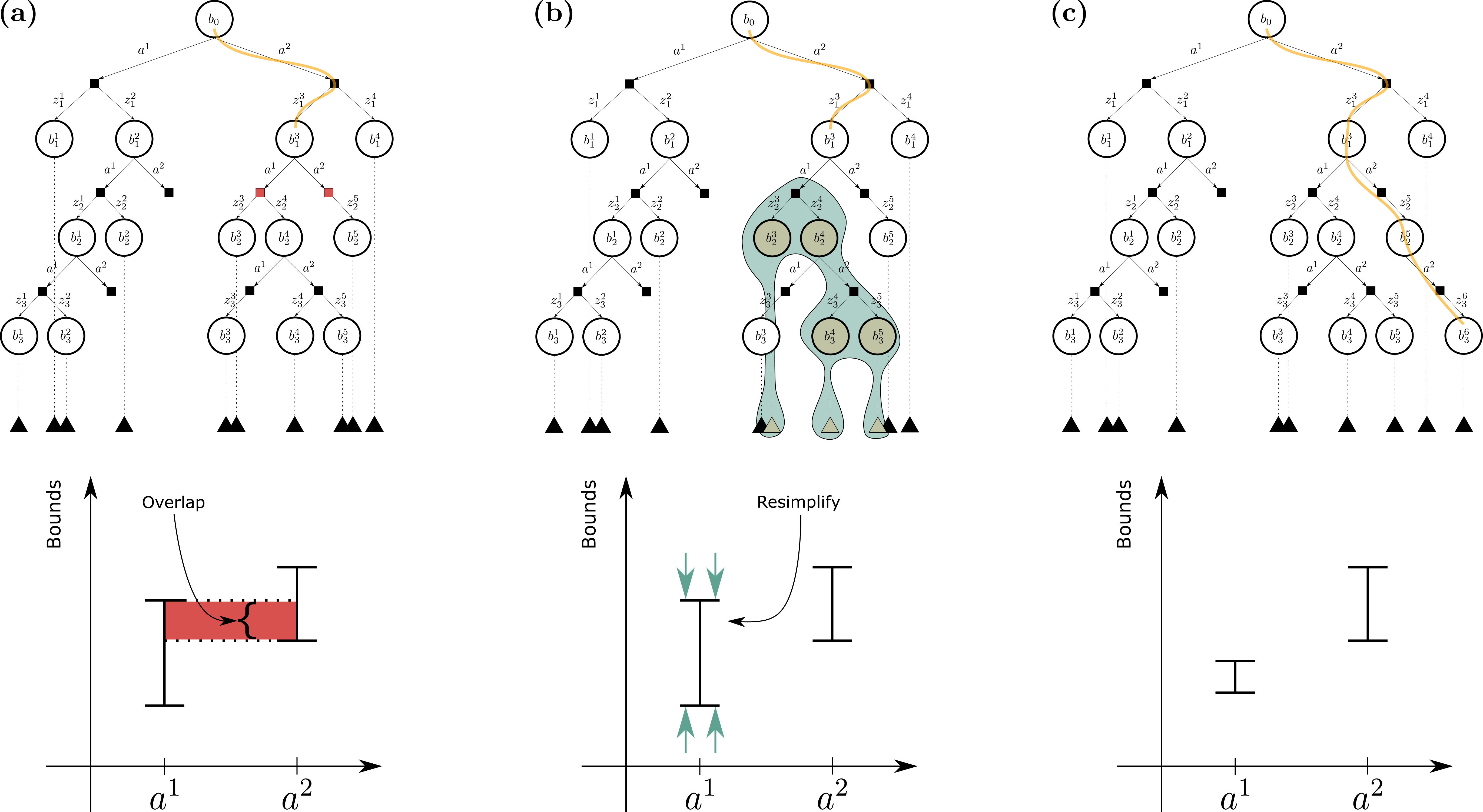}
	\caption{Illustration of our approach. The circles denote the belief nodes, and the rectangles represent the belief-action nodes. Rollouts, emanating from each belief node, are indicated by dashed lines finalized with triangles.  \textbf{(a)} The simulation starts from the root of the tree, but at node $b_1^3$ it can not continue due to an overlap of the child nodes (colored red) bounds. \textbf{(b)} One of the red colored belief-action nodes is chosen, and resimplification is triggered from it down the tree to the leaves (shaded green area in the tree). The beliefs and rollouts inside the green area (colored by light brown)  undergo resimplification if decided so. This procedure results in tighter bounds. \textbf{(c)} After the bounds got tighter, nothing prevents the SITH-PFT from continuing down from node $b_1^3$ guaranteeing the Tree Consistency. If needed, additional resimplifications can be commenced.}  
	\label{fig:resimpl_illustration}
	\vskip -0.2in
\end{figure}
POMDPs have proven to be a celebrated mathematical framework for planning under uncertainty [\cite{Kurniawati08rss, Silver10nips, Ye17jair, Sunberg18icaps, Garg19rss}]. During the planning session, an agent is given some goal and attempts to find the optimal action to execute. Since the agent operates under uncertainty, it maintains a \emph{belief} over the state and reasons about its evolution while planning.  The standard way to represent a general belief distribution is by a set of weighted particles. In a finite horizon setting, the agent performs planning with a fixed number of steps ahead of time. Equipped with motion and observation models, the agent has to consider every possible realization of the future observations for every available reactive action sequence (policy) of the length of the horizon. In a sampled form, this abundance of possible realizations of action observation pairs constitutes a \emph{belief tree}.  Building the full belief tree is intractable since each node in the tree repeatedly branches with all possible actions and all possible observations. The number of nodes grows exponentially with the horizon.
Additionally, the number of possible states grows exponentially with the state space dimension, and consequently, an adequate representation of the belief requires more particles. Those last two problems are known as the \emph{curse of history} and the \emph{curse of dimensionality} respectively. MCTS based algorithms tackle those problems by (a) building the belief tree incrementally and revealing only the ``promising''  parts of the tree, and (b) representing the belief as a fixed size set of weighted state samples (particles). 
An inherent part of MCTS based algorithms is the Upper Confidence Bound (UCB) technique \cite{Kocsis06ecml} designed to balance exploration and exploitation while building the belief tree. This technique assumes that calculating the reward over the belief node does not pose any computational difficulty. Information-theoretic rewards violate this assumption.
\subsection{Related Work} 
Incorporation of information-theoretic reward into POMDP is a long standing effort. Earlier attempts such as \cite{Dressel17icaps} were tackling offline solvers. 
Monte Carlo Tree Search made a significant breakthrough in overcoming the course of history. However, when the reward is a general function of the belief, the origin of the computational burden is shifted towards the reward calculation. Moreover, belief-dependent reward prescribes the complete set of belief particles at each node in the belief tree.   Therefore, algorithms such as POMCP  \cite{Silver10nips}, and its numerous predecessors are inapplicable since they simulate each time a single particle down the tree when expanding it.  DESPOT based algorithms behave similarly \cite{Ye17jair}, with the DESPOT-$\alpha$ as an exception \cite{Garg19rss}. 
DESPOT-$\alpha$ simulates a complete set of particles. However, this algorithm depends on $\alpha$-vectors. In particular, as in other DESPOT-like algorithms, the belief tree is determinized. Therefore, sibling belief nodes have identical particles and are distinct solely by the weights. DESPOT-$\alpha$ leverages this regard and uses the $\alpha$-vectors to efficiently approximate the lower bound of the value function of the sibling belief nodes without expanding them. Since DESPOT-like methods are based on gap heuristic search \cite{Kochenderfer22book}, this lower bound is an essential part of the exploration strategy. In other words, the DESPOT-$\alpha$ tree is built using  $\alpha$-vectors, such that they are an indispensable part of the algorithm. Note that an integral part of this approach is that the reward is state dependent, and the reward over the belief is merely expectation over the state reward. DESPOT-$\alpha$ does not support belief-dependent rewards since it contradicts the application of the $\alpha$-vectors. 
The only approach posing no restrictions on the structure of belief-dependent reward and not suffering from limiting assumptions is 
Particle Filter Tree (PFT). The idea behind  PFT is to apply MCTS over Belief-MDP. \cite{Sunberg18icaps} augmented PFT  with Double Progressive Widening and coined the name PFT-DPW. PFT-DPW utilizes the UCB strategy and maintains a complete belief particle set at each belief tree node.  
Recently, \cite{Fischer20icml} presented Information Particle Filter Tree (IPFT), a method to incorporate  information-theoretic rewards into PFT. The IPFT planner is remarkably fast. It simulates small subsets of particles sampled from the root of the belief tree and averages entropies calculated over these subsets. However,  differential entropy estimated from a small-sized particle set can be significantly biased. This bias is unpredictable and unbounded, therefore, severely impairs the performance of the algorithm. In other words, celerity comes at the expense of quality. Often times the policy defined by this algorithm is suboptimal. 
\citet{Fischer20icml} provide guarantees solely for the asymptotic case, i.e, the number of state samples (particles) tends to infinity. Asymptotically their algorithm behaves precisely as the PFT-DPW in terms of running speed and performance. Yet, in practice the performance of IPFT in terms of optimality can degrade severely compared to PFT-DPW. Moreover, \cite{Fischer20icml} does not provide any reliable study of  comparison of IPFT against PFT-DPW with an information-theoretic reward. Prompted by this insight, we chose the  PFT-DPW as our \emph{baseline} approach, which we aim to accelerate. In contrast to IPFT, our approach 
explicitly guarantees an \emph{identical} solution to PFT-DPW with information-theoretic reward, for \emph{any} size of particle set representing the belief and serving as input to PFT-DPW.

The computational burden incurred by the complexity of POMDP planning inspired many research works to focus on approximations of the problem, e.g., \cite{Hoerger19isrr}. Typically, approximation based planners show asymptotical guarantees, e.g., the convergence of the algorithms. Recently, the novel paradigm of simplification has appeared in literature \cite{Zhitnikov21arxiv, Sztyglic21arxiv, Elimelech18ijrr_submitted}. The simplification is concerned with carefully replacing the nonessential elements of the decision making problem and quantifying the impact of this relaxation. Specifically, simplification methods are accompanied by stringent guarantees.
\subsection{Contribution}
We provide a novel algorithmic framework based on converging bounds on a belief-dependent reward. Our method is guaranteed to yield the same action and belief tree as the most general algorithm suitable for such belief-dependent rewards (PFT-DPW). The proposed technique is applicable with any converging to the reward bounds.  In this paper, we focus on information-theoretic rewards, in particular, differential entropy. 
\section{Background}


\subsection{POMDPs with belief-dependent rewards} \label{sec:pomdp_bg}
POMDP is a tuple  $	\langle \mathcal{X}, \mathcal{A}, \mathcal{Z}, T, O, \rho, \gamma, b_{0}\rangle $
where $\mathcal{X}, \mathcal{A}, \mathcal{Z}$ are state, action, and observation spaces with $x \in \mathcal{X}, a \in \mathcal{A}, z \in \mathcal{Z}$ the momentary state, action, and observation, respectively, $T(x,a,x')= \mot{a}{x}{x'} $ is the stochastic transition model from the past state $x$ to the subsequent  $x'$ through action $a$, $O(z,x)=\observ{x}{z}$ is the stochastic observation model, $\gamma \in [0,1]$ is the discount factor, $b_0$ is the belief over the initial state (prior), and $\rho$ is the reward operator. Let $\hist$ denote \emph{history} of actions and observations obtained by the agent up to time instance $k$ and the prior belief. The posterior belief is given by $b_k(x_k)= \probd(x_k \sep h_k)$. The \emph{policy} is a mapping from belief to action spaces $a_k=\pi_k(b_k)$. The policy for $L$ consecutive steps ahead is denoted by $\pi \bydef \pi_{k:k+L-1}$. The decision making goal is to find the optimal policy $\pi^*$ maximizing 
\begin{align}
	V^{\pi}(b_k) = \expt{z_{k+1:k+L}}{\sum_{i=k}^{k+L-1}\gamma^{i-k}\rho(b_i, \pi_i(b_i), z_{i+1}, b_{i+1})\sep \pi} \ \ \text{s.t.} \ \ b_{i+1} = \psi(b_i, \pi_i(b_i), z_{i+1}), \label{eq:value}
\end{align}
where $\psi$ is the Bayesian belief update method. 
Bellman form  representation of \eqref{eq:value} is  $V^{\pi^*}(b) = \maxim{\argmax}{a_k} \ Q(b_k, a_k)$, where 
\begin{align}
Q(b_k, a_k) = \expt{z_{k+1}}{\rho(b_k, a_k, z_{k+1}, b_{k+1})\sep b_k, a_k}  + \expt{z_{k+1}}{V^{\pi^*}(\psi(b_k, a_k, z_{k+1})\sep b_k, a_k}. \label{eq:qbellman}
\end{align}
In our generalized formulation, the reward 
$\rho(b_k, a_k, z_{k+1}, b_{k+1}) = r^{x}(b_k, a_k)+\lambda r^I(b_k, a_k, z_{k+1}, b_{k+1})$   is a function of two subsequent beliefs, an action and an observation. Specifically, our reward is 
\begin{align}
&\rho(b_k, a_k, z_{k+1}, b_{k+1}) = \expt{x_k \sim b_k}{r(x_k, a_k)} - \lambda \hat{\mathcal{H}}(b_{k}, a_k, z_{k+1}, b_{k+1}), \label{eq:rew} 
\end{align}
where $r(x_k,a_k)$ is state and action dependent reward, and $r^{x}(b_k, a_k)$ is the expectation with regard to the state. $r^I(b_k, a_k, z_{k+1}, b_{k+1})$ is an information-theoretic reward, which in general can be dependent on consecutive beliefs and the elements relating them (e.g.~information gain). $-\hat{\mathcal{H}}(b_{k}, a_k, z_{k+1}, b_{k+1})$ is an estimator of our information-theoretic reward weighted by $\lambda$. Yet, since such estimators do not commonly have a closed-form expression for non-parametric beliefs represented by a set of samples, one has to consider an estimator $\hat{\mathcal{H}}$ of $\mathcal{H}$ (e.g., \cite{Boers10fusion}).  As shall be seen, our chosen estimator requires also previous belief $b_{k}$, chosen action $a_k$, and received observation $z_{k+1}$. Depending on the estimation method, the inputs can vary. Using the structure of~\eqref{eq:rew}, 
%
\begin{align}
Q(b_k, a_k) = Q^{x}(b_k, a_k) + \lambda Q^{I}(b_k, a_k), \label{eq:dissectQ}
\end{align}
where $Q^{x}$ is induced by state dependent rewards and $Q^{I}$ by the information-theoretic rewards. They are constituted by $L$ elements of the form  $\expt{x_i \sim b_i}{r(x_i, a_i)}$ and $-\entropy{b_{i}, a_i, z_{i+1}, b_{i+1}}$, respectively. The $Q^{x}$ element is easy to calculate, thus out of our focus, whereas the  $Q^{I}$ is computationally expensive to compute. From here on, for the sake of clarity, we will use the notation $h$ and $b$ interchangeably. 

\subsection{MCTS over Belief-MDP (PFT)}
In this section, we outline the UCB based MCTS over Belief-MDP.  
The algorithm constructs the policy tree by executing multiple simulations. Each simulation adds a single belief node to the belief tree or terminates by terminal state or action. To steer towards more deep and beneficial simulations, MCTS chooses action $a^{\dagger}$ at each belief node according to following rule 
\begin{align}
&a^{\dagger} = \maxim{\argmax }{a \in \mathcal{A}}  \ \text{UCB} (ha)\quad \text{UCB}(ha) = Q(ha) + c\cdot \sqrt{\frac{\log(N(h))}{N(ha)}},  \label{eq:ucb}
\end{align}
where $N(h)$ is the visitation count of belief node defined by the history $h$, $N(ha)$ is the visitation count of the belief-action node, $c$ is the exploration parameter and,  $Q(ha)$ is the  approximation of the belief-action value function $Q$ for node $ha$ obtained by simulations.  
When the action is selected, a question arises either to open a new branch in terms of observation and posterior belief or to continue through one of the existing branches. In continuous spaces, this is resolved by the Progressive Widening technique \cite{Sunberg18icaps}.  If a new branch is expanded, an observation $o$ is created from state $x$ drawn from the belief $b$. 


\section{Approach}
SITH-PFT (Alg.~\ref{alg:simp_planning}) follows the same algorithmic baseline as PFT.  We adhere to the conventional notations \cite{Sunberg18icaps} and denote by $G_{\text{PF}(m)}(bao)$ a generative model receiving as input the belief $b$, an action $a$ and an observation $o$, and producing the posterior belief $b'$ and  the mean reward over the state $r^x(b, a)$. For belief update, we use a particle filter based on $m$ belief samples.   
Instead of calculating the immediate information-theoretic rewards and the corresponding $Q^I$ function estimates, we calculate low-cost lower and upper bounds $\ell, u$  over the information-theoretic rewards and corresponding bounds $\lb$, $\ub$ over the $Q^I$ function. These bounds are adaptive and can be tightened on demand. We call the process of tightening  ``resimplification''. 
\begin{algorithm}[h]
		\caption{SITH-PFT}
		\begin{algorithmic}[1]
			\Procedure{Plan}{belief: $b$}
			\State $h \leftarrow \emptyset$ 
			\For{$i \in 1:n$} 
			\State	\Call{Simulate}{$b$, $d_{\text{max}}$, $h$}
			\EndFor
			\State \Return \Call{Action Selection}{$b$, $h$} \Comment{called with nullified exploration constant $c$}
			\EndProcedure
			\Procedure{Simulate}{belief: $b$, depth: $d$, history: $h$}
			\If { $d=0$ }
			\State \Return $0$
			\EndIf
			\State $a \leftarrow $ \Call{Action Selection}{$b$, $h$}
			\If {$|C(ha)| \leq k_o N(ha)^{\alpha_{o}}$} 
			\State $o  \leftarrow$ sample $x$ from $b$, generate $o$ from $(x,a)$
			\State $b' , r^x \leftarrow G_{\text{PF}(m)}(bao)$ 
			\State Calculate initial $u', \ell'$ for $b'$ based on $s\leftarrow 1$ \Comment{minimal simp. level}
			\State $C(ha) \leftarrow C(ha) \cup \{(r^x, \ell', u', b', o)\} $ 
			
			\State $R, L, U \leftarrow r^x, \ell', u' + \gamma$ \Call{Rollout}{$b'$, $hao$, $d-1$}
			\Else 
			\State $(r^x, \ell', u', b', o) \leftarrow $ sample uniformly from $C(ha)$
			\State $R, L, U \leftarrow r^x, \ell', u' + \gamma$ \Call{Simulate}{$b'$, $hao$, $d-1$}
			\EndIf
			\If {deepest resimplification depth $< d$ } \Comment{accounting for updated deeper in the tree bounds. See section~\ref{sec:bounds_recon}}
			\State reconstruct $\lb(ha), \ub(ha)$ 
			\EndIf 
			\State $N(h) \leftarrow N(h)+1$
			\State $N(ha) \leftarrow N(ha)+1$
			\State $Q^x(ha) \leftarrow Q^x(ha) + \frac{R-Q^x(ha)}{N(ha)}$ 
			\State $\lb(ha) \leftarrow \lb(ha) + \frac{L-\lb(ha)}{N(ha)}$
			\State $\ub(ha) \leftarrow \ub(ha) + \frac{U-\ub(ha)}{N(ha)}$
			\State \Return $R, L, U$
			\EndProcedure
		\end{algorithmic}
		\label{alg:simp_planning}
	\end{algorithm}	
In turn, these bounds induce bounds over UCB.  As we discuss in detail next, an essential aspect of our approach is using these bounds to achieve the exact same action selection as UCB without exactly calculating the $Q$ function and UCB. To this end we present a novel action selection method (Alg.~\ref{alg:action_dis}). Crucially, by tightening the bounds only to a minimal needed extent (Alg.~\ref{alg:resimpl}), we guarantee the same tree connectivity and calculated optimal action compared to PFT-DPW, but faster. 
We devote the subsequent section to the bounds and explain how they pertain to SITH-PFT.

\subsection{Information theoretic bounds} \label{sec:bounds}
In the setting of continuous state space and nonparametric belief represented by $m$ weighted particles $b \bydef \{w^i, x^i\}^m_{i=1}$, the estimation of differential entropy is not a simple task. Typically, such estimators' complexity is squared in the number of particles \cite{Fischer20icml, Boers10fusion}. We use \cite{Boers10fusion} as a reward function and utilize
the bounds over it, developed by \cite{Sztyglic21arxiv}. The bounds can be tightened on demand incrementally without an overhead. Namely, after a few bounds-tightening iterations they are just the reward itself and the entire calculation is time-equivalent to calculating the original reward.  We define the bounds over the minus differential entropy estimator for $b_{k+1}$ as (see supplementary~\ref{sec:supp_bounds} for the full terms) 
\begin{equation}
\ell(b_{k}, a_k, z_{k+1}, b_{k+1}; A^s_k, A^s_{k+1})\! \leq \! -\hat{\mathcal{H}}(b_{k}, a_k, z_{k+1}, b_{k+1}) \! \leq \! u(b_{k}, a_k, z_{k+1}, b_{k+1}; A^s_k, A^s_{k+1}), \label{eq:bounds}
\end{equation}
where $s$ is the discrete level of simplification $s \in \{1, 2, \ldots, M\}$. Higher levels of simplification correspond to tighter, and lower levels of simplification correspond to looser bounds. $A^s_k$, $A^s_{k+1}$ are the simplification level corresponding sets of indices. Specifically, $b_{k}, b_{k+1}$ are each represented as a set of $m$ weighted particles. We keep track over the indices of particles that were chosen for the bounds calculation. Namely,  $A^s\subseteq \{1,2,\ldots,m\}$ and $|A^s|= m^s$. Each subsequent level (low to high) defines a larger set of indices. 
Sometimes the bounds are not close enough to select the same action as UCB. In this case, our modified action selection routine triggers the resimplification process.  When resemplification is carried out, new indices are drawn from the sets $\{1,2,\ldots,m\} \setminus A^s_k$ and $\{1,2,\ldots,m\} \setminus A^s_{k+1}$  respectively, and added to the sets  $A^s_k$ and $A^s_{k+1}$. This operation promotes the simplification level to $s+1$ and defines  $A^{s+1}_k$ and $A^{s+1}_{k+1}$. Importantly, increasing simplification level is done incrementally (as introduced by \cite{Sztyglic21arxiv}). Thus, when we refine the bounds $\ell, u$ (Alg~\ref{alg:resimpl} lines 3,12,18), from simplification level $s=1$ all the way to $s=M$  (worst case scenario) the time complexity  is equivalent to calculation of $\hat{\mathcal{H}}(\cdot)$. When $s=M$ , it holds that $\ell(\cdot) = -\hat{\mathcal{H}}(\cdot) = u(\cdot)$. Importantly, by caching the shared calculations of the two bounds,
we never repeat the calculation of these values and obtain maximal speedup. 
The immediate bounds \eqref{eq:bounds} induce bounds over $Q^{I}(\cdot)$. In MCTS, the $Q$ approximation is a mean over simulations. Each simulation yields  a sum of discounted cumulative rewards. Therefore, if we replace the reward $-\hat{\mathcal{H}}(\cdot)$ with the bounds from \eqref{eq:bounds} we will get corresponding discounted cumulative upper and lower bounds. Averaging the simulations, in the same manner (Alg.~\ref{alg:simp_planning} lines 29-30), yields
\begin{align}
\lb(\cdot) \leq Q^I(\cdot) \leq \ub(\cdot). \label{eq:Qibounds}
\end{align}
\subsection{UCB bounds}\label{sec:ucb_bounds}
Since the MCTS tree is built upon \eqref{eq:ucb}, using \eqref{eq:dissectQ} and \eqref{eq:Qibounds} we denote  UCB upper and lower bounds as
\begin{align}
&\lucb(ha) \triangleq Q^x(ha) + \lambda \lb(ha) + c\cdot \sqrt{\frac{\log(N(h))}{N(ha)}},  \\
&\uucb(ha) \triangleq Q^x(ha) + \lambda \ub(ha) + c\cdot \sqrt{\frac{\log(N(h))}{N(ha)}}.
\end{align}
\subsection{Guaranteed belief tree consistency}\label{sec:consistency}
\begin{algorithm}[t]
	\caption{Action Selection}
		\begin{algorithmic}[1]
			\Procedure{Action Selection}{$b$, $h$}
			\While{true}
			\State Status, $a \leftarrow$ \Call{Select Best}{$b$, $h$}
			\If{Status}
			\State break
			\Else
			\ForAll {$b', o \in C(ha)$} 
			\State \Call{Resimplify}{$b'$, $hao$}
			\EndFor
			\State reconstruct $\lb(ha), \ub(ha)$ 
			\EndIf
			\EndWhile
			\State return a				
			\EndProcedure
			
			\Procedure{Select Best}{$b$, $h$}
			\State Status $\leftarrow$ true
			\State $\tilde{a} \leftarrow \maxim{\argmax}{a}\{\lucb(ha)\}$ 
			\State gap $\leftarrow 0$ 
			\State child-to-resimplify $\leftarrow \tilde{a}$
			\ForAll{$ha$ children of $b$}
			\If{$\lucb(h\tilde{a}) < \uucb(ha)  \land a \neq \tilde{a}$ } 
			\State Status $\leftarrow$ false
			\If {$\ub(ha) - \lb(ha) >$ gap}
			\State gap $\leftarrow \ub(ha) - \lb(ha) $
			\State child-to-resimplify $\leftarrow$ $a$ 
			\EndIf
			\EndIf
			\EndFor
			\State \textbf{return} Status, child-to-resimplify
			\EndProcedure
		\end{algorithmic}
		\label{alg:action_dis}
\end{algorithm}
In this section, we define the Tree Consistency and explain and prove the equivalence of our algorithm to PFT-DPW. 
\begin{defn}[Tree consistent algorithms]\label{def:tree_consis}
Consider two algorithms, constructing a belief tree. Assume every common sampling operation for the two algorithms uses the same seed. 
The two algorithms are \emph{tree consistent} if two belief trees constructed by the algorithms are identical in terms of actions, observations, and visitation counts. 	
\end{defn}
Our approach leans on a specific action selection procedure inside the tree, which differs from the PFT.  At every belief node we mark as a candidate action the one that maximizes the lower bound $\lucb$ as such 
\begin{align}
\tilde{a} = \maxim{\argmax }{a \in \mathcal{A}}  \ \lucb(ha).
\end{align}
If  $\forall a \neq \tilde{a}$, $\lucb(h\tilde{a}) \geq \uucb(ha)$, 
there is no overlap (Fig.~\ref{fig:resimpl_illustration} \textbf{(c)}) and we can announce $\tilde{a}$  is identical to $a^{\dagger}$, i.e.,~the action that would be returned by PFT using \eqref{eq:ucb} and the tree consistency was not compromised.  Else, the bounds need to be tightened, so we may guarantee the tree consistency. We examine the $ha$ siblings of $h\tilde{a}$, fulfilling   $ a \neq \tilde{a}:\lucb(h\tilde{a}) < \uucb(ha)$ (Fig.~\ref{fig:resimpl_illustration} \textbf{(a)}). Our next step is to tighten the bounds via resimplification (Fig.~\ref{fig:resimpl_illustration} \textbf{(b)}) until there is no overlap. 
When some sibling nodes  have overlapping bounds, we strive to avoid tightening all of them at once since fewer resimplifications lead to a greater speedup. 
\begin{algorithm}[t]
		\caption{Resimplification}
		\begin{algorithmic}[1]
			\Procedure{Resimplify}{$b$, $h$}
			\If { $b$ is a leaf}
			\State \Call{Refine$_{\{\ell,u\}}$}{$b$}
			\State \Call{Resimplify Rollout}{$b$, $h$}
			\State \textbf{return}
			\EndIf
			\State $\tilde{a} \leftarrow \maxim{\argmax}{a} \{N(ha)\cdot(\ub(ha)-\lb(ha))\}$ 
			\ForAll {$b', o \in C(h\tilde{a})$}
			\State \Call{Resimplify}{$b'$, $h\tilde{a}o$}
			\EndFor
			\State reconstruct $\lb(h\tilde{a}), \ub(h\tilde{a})$
			\State \Call{Refine$_{\{\ell,u\}}$}{$b$}
			\State \Call{Resimplify Rollout}{$b$, $h$}
			\State \textbf{return} 
			\EndProcedure
			
			\Procedure{Resimplify Rollout}{$b$, $h$}
			\State $b^{\text{rollout}} \leftarrow$ find weakest link in rollout
			\State \Call{Refine$_{\{\ell,u\}}$}{$b^{\text{rollout}}$}
			\EndProcedure
			
			\Procedure{Refine$_{\{\ell,u\}}$}{$b$}
			\State if \eqref{eq:cond} holds for $b$, refine its $\ell, u$ and promote its simplification level
			\EndProcedure
		\end{algorithmic}
		\label{alg:resimpl}
\end{algorithm}
Thus, among them we pick a single $ha$ node that induces the biggest ``gap'', denoted by $g$, between its bounds (see Alg.~\ref{alg:action_dis} lines 20-28), where
\begin{align}
g(ha) \triangleq \ub(ha)-\lb(ha). \label{eq:gap}
\end{align}
Further, we tighten the bounds down the branch of the chosen node (see Alg.~\ref{alg:action_dis} lines 7-9) for every member of $C(ha)$, the set of children of $ha$.  Since the bounds converge to the actual information reward we can guarantee the algorithm will pick a single action after a finite number of ``bounds-tightening'' iterations (resimplification); thus, tree consistency is assured. In the following section, we delve into the resimplification procedure.
\subsection{Resimplification}\label{sec:resimplification}
In this section, we explain how resimplification is done. The algorithmic scheme is formulated in a general manner. However, it is guided by a specific strategy meant to minimize the number of times we tighten the bounds (as mentioned in Sec.~\ref{sec:consistency}). We denote this strategy as \textit{Resimplification Strategy}. We assume this strategy satisfies two conditions to guarantee tree consistency.

\begin{assumption}[Convergence]\label{as:converg_resimpl}
	When using a \emph{converging strategy}, each call to resimplify on the children of $ha$, tightens the $\lucb(ha), \uucb(ha)$ bounds (unless they are already equal).
\end{assumption}
\begin{assumption}[Finite-time]\label{as:finite_resimpl}
	When using a \emph{finite-time strategy}, after a finite number of calls to resimplify on the children of $ha$, it holds $\lucb(ha)=\text{UCB}(ha)=\uucb(ha)$.
\end{assumption}

\subsubsection{Resimplification algorithmic scheme}

Consider a belief-action node $ha$ at level $d$ with  $\lb(ha), \ub(ha)$. Assume the algorithm chooses it for bounds tightening, as described in Sec.~\ref{sec:consistency} and Alg.~\ref{alg:action_dis} line 3. All tree nodes that $ha$ is an ancestor to them, contribute their immediate $\ell, u$ bounds to $\lb(ha), \ub(ha)$ calculation. Thus, to tighten $\lb(ha), \ub(ha)$, we can potentially choose any candidate nodes in the subtree of $ha$.   Every child belief node of $ha$ is sent to the resimplification routine (Alg.~\ref{alg:action_dis} lines 7-9), which performs four tasks. 
Firstly, it chooses the action (Alg~\ref{alg:resimpl} line 7) that will participate in the subsequent resimplification call and sends all its children beliefs nodes to the recursive call down the tree (Alg.~\ref{alg:resimpl} line 8-10).  Secondly, it refines the belief node  $\ell, u$ according to the specific \emph{resimplification strategy} (Alg~\ref{alg:resimpl} lines 3,12,18). Thirdly, it reconstructs $\lb(ha)$, $\ub(ha)$ once all the children belief nodes of $ha$ have returned from the resimplification routine (Alg~\ref{alg:resimpl} line 11).  
Fourthly, it engages the rollout resimplification routine according to the specific \emph{resimplification strategy} (Alg~\ref{alg:resimpl} lines 4, 13). Upon completion of this resimplification call initiated at $ha$, we get tighter immediate bounds of some of $ha$ descendant belief nodes (including rollouts nodes). Accordingly, all of $ha$ descendant belief-action nodes bounds ($\lb, \ub$) were updated.

\subsubsection{Specific resimplification strategy}\label{sec:chosen_startegy}
Specifically, we decide to refine $\ell', u'$ of a belief node $h'$ with depth $d'$ if 
\begin{align}
\gamma^{d-d'} \cdot (u'- \ell') > \frac{1}{d} g(ha), \label{eq:cond}
\end{align}
where
$g(ha)$ corresponds to the gap \eqref{eq:gap} of the belief-action node $ha$ that initially triggered resimplifcation in Alg.~\ref{alg:action_dis} line 24. 
The explanation to \eqref{eq:cond} resimplification strategy is rather simple. The right hand side of \eqref{eq:cond} is the mean gap per depth/level in the sub-tree with $ha$ as its root and spreading downwards to the leaves.  Naturally, some of the nodes in this subtree have $u-\ell$  above the mean gap, and some under. We wish to locate and refine all the ones above. For the left-hand side of  \eqref{eq:cond};  the rewards are accumulated and discounted according to their depth. Thus, when comparing $ha$ node with depth $d$ to belief node $h'$ with depth $d'$, we must account for the relative proper discount factor. Note the depth identified with the root is $d_{\max}$ as seen in Alg.~\ref{alg:simp_planning} line 4, and the leafs are distinguished by depth $d=0$.
\begin{figure}[t]
	\centering
	\begin{subfigure}{0.33\textwidth}            
		\includegraphics[width=\textwidth]{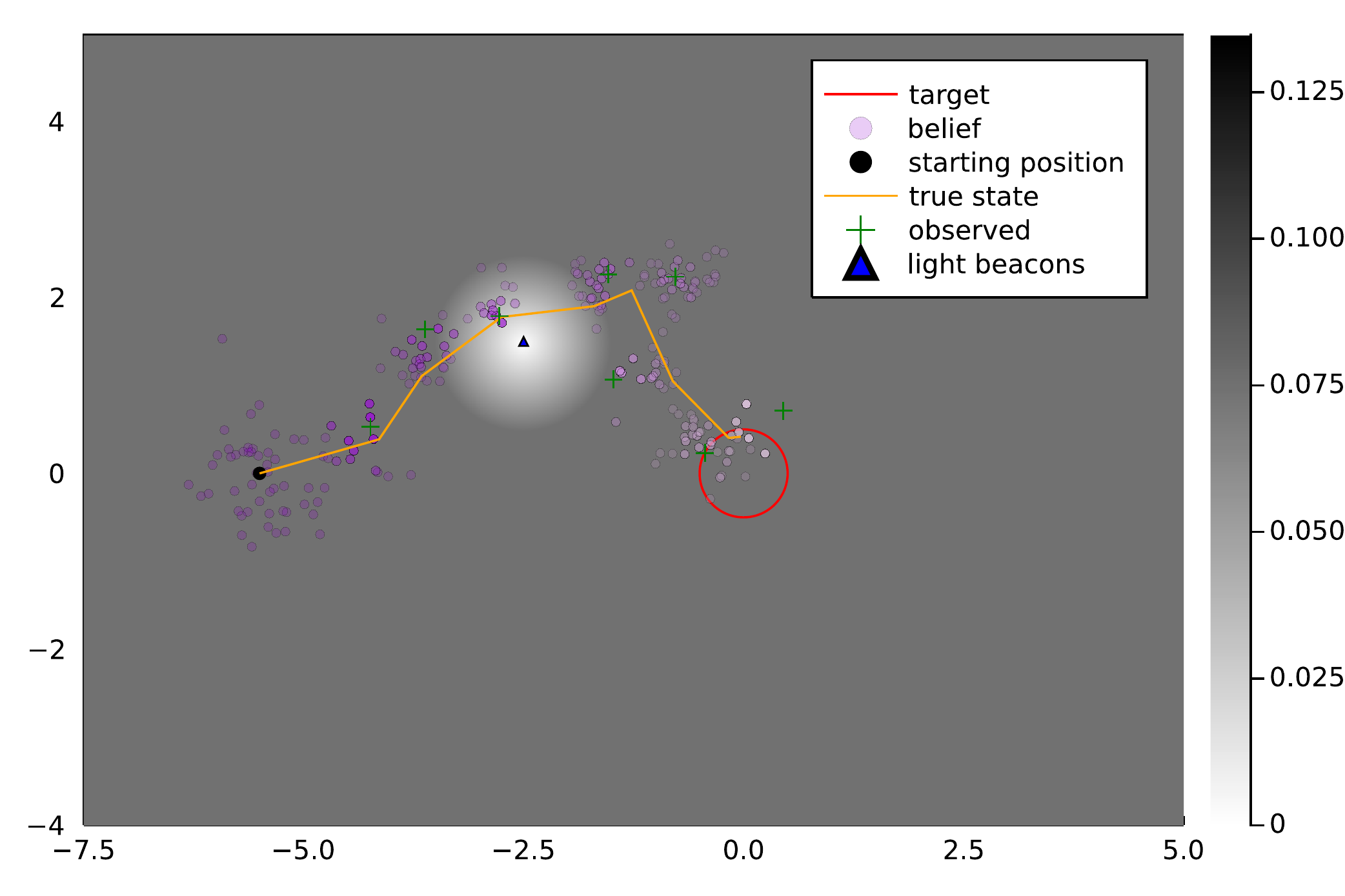}
		\caption{SITH-PFT}
		\label{fig:sith-pft_cl2d}
	\end{subfigure}%
	\begin{subfigure}{0.33\textwidth}
		\centering
		\includegraphics[width=\textwidth]{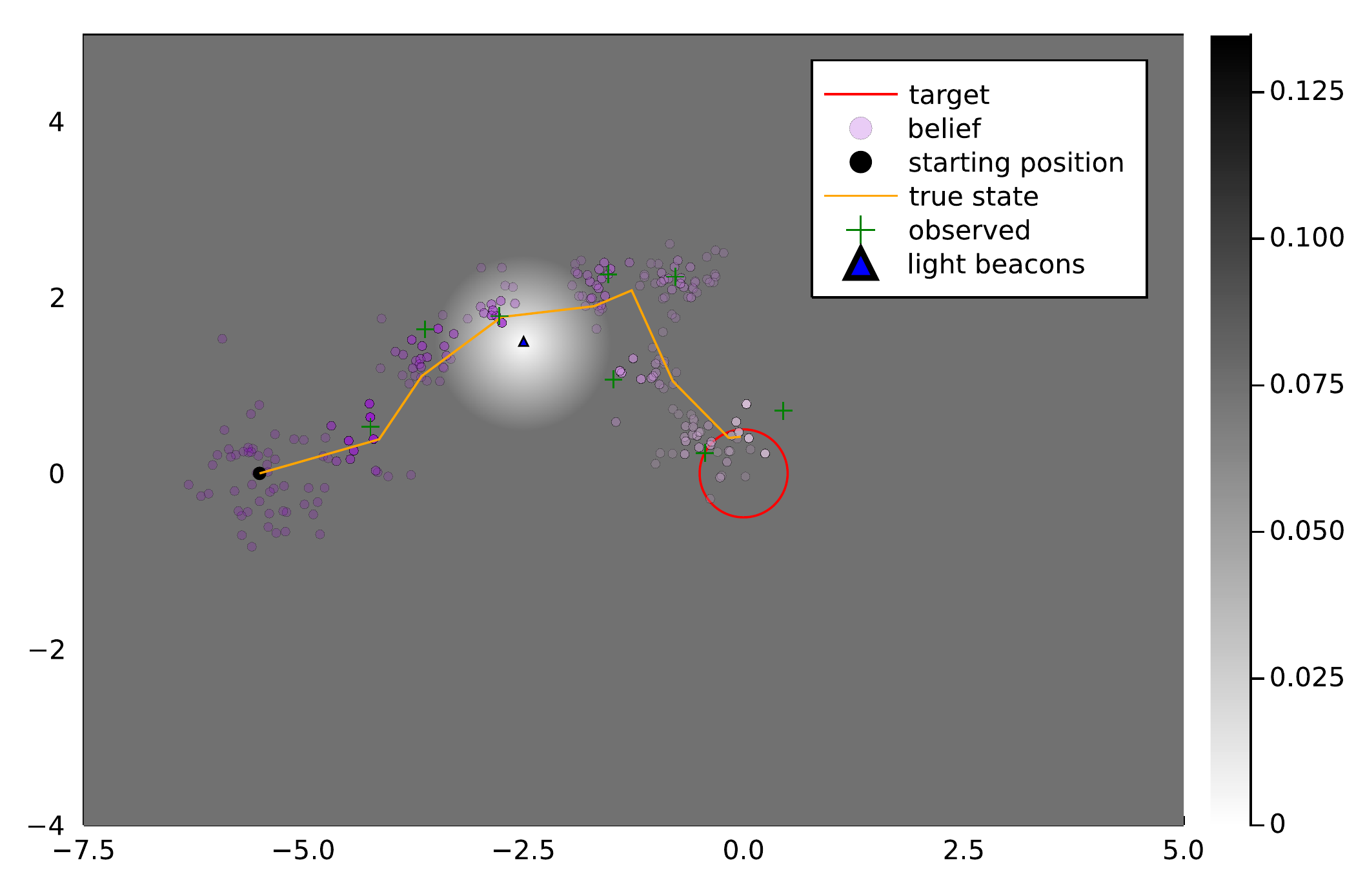}
		\caption{PFT-DPW}
		\label{fig:pft_cl2d}
	\end{subfigure}
	\begin{subfigure}{0.33\textwidth}
		\centering
		\includegraphics[width=\textwidth]{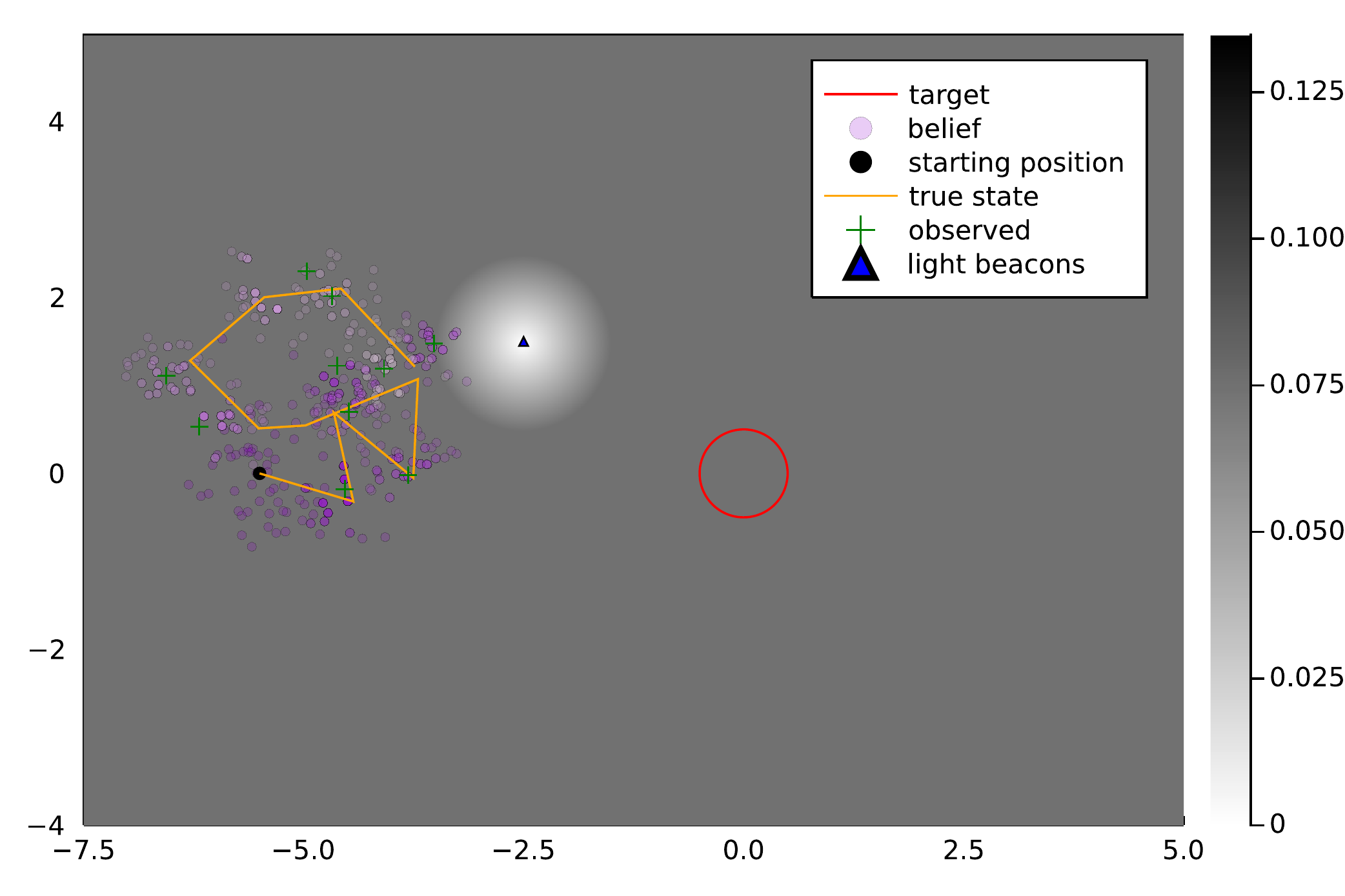}
		\caption{IPFT}
		\label{fig:ipft_cl2d}
	\end{subfigure}
	\caption{2D Continuous Light Dark. The agent starts from an initial unknown location and is given an initial belief. The goal is to get to location $(0,0)$ (circled in red) and execute the terminal action. Near the beacon (white light) the observations are less noisy. We consider multi-objective function, accounting for the distance to the goal and the differential entropy approximation (with the minus sign for reward notation). Executing the terminal action inside the red circle gives the agent a large positive reward but executing it outside it, will yield a large negative reward.}
	\label{fig:cl2d}
	\vskip -0.2in
\end{figure}
For each rollout originating from the tree belief node, we find the rollout node with the biggest $u-\ell$ fulfilling \eqref{eq:cond} term locally in the rollout and resimplify it (Alg~\ref{alg:resimpl} lines 4,13).
To choose the action to continue resimplifcaiton down the tree, we take the action corresponding to the belief-action node with the largest gap weighted by its visitation count (Alg~\ref{alg:resimpl} line 7). With this strategy, we aim to leave the belief tree at the lowest possible simplification levels whilst still guarantee tree consistency.
\subsubsection{Reconstructing the bounds} \label{sec:bounds_recon}
If the action selection procedure triggered a resimplification, it modified the bounds through the tree. Since the resimplifcation works recursively, it reconstructs the belief-action node bounds coming back from the recursion (Alg.~\ref{alg:resimpl} line 11).  Similarly, the action dismissing procedure reconstructs $\lb$, and $\ub$ of the belief-action node at which the action dismissing is performed (Alg.~\ref{alg:action_dis} line 10).  Moreover, on the way back from the simulation, we shall update the ancestral belief-action nodes of the tree. Specifically, we are required to reconstruct each $\lb$ and $\ub$  higher than the deepest starting point of the resimplification  (Alg.~\ref{alg:simp_planning} line 23-25). 
Reconstruction is essentially a double loop. To reconstruct  $\ub(ha),\lb(ha)$ we first query for all belief children nodes $hao$. We then query all belief-action nodes that are children to the $hao$, i.e., $haoa'$. The possibly modified immediate bounds $\ell$ and $ u$ are taken from $hao$ nodes and the $\ub(\cdot)$, $\lb(\cdot)$ bounds are taken from the  $haoa'$ nodes.  Importantly, each of the bounds is weighted according to the proper visitation count. 
\subsection{Guarantees} \label{sec:guarantees}
Assuming a \emph{converging and finite-time resimplification strategy}, the following theorems are satisfied:
\begin{thm}\label{thm:tree_consis}  
	The SITH-PFT and PFT are Tree Consistent Algorithms. 
\end{thm}
\begin{thm}\label{thm:sol_consis}
	The SITH-PFT provides the same solution as PFT.
\end{thm}
\begin{thm}\label{thm:valid_strategy}
	The specific resimplificaiton strategy from Sec.~\ref{sec:chosen_startegy} is a converging and finite-time resimplification strategy.
\end{thm}
See full proofs of the theorems and time complexity analysis using the specific bounds in the supplementary~\ref{sec:supp_proofs},\ref{sec:time_complexity}.
Note other resimplification strategies are possible, see supplementary~\ref{sec:supp_more_strategies}. 

%
\section{Experiments}
\label{sec:exp}
In the continuous setting with information-theoretic rewards, many common POMDP benchmarks (e.g., rock sampling, laser tag) are inadequate. We turn to the challenging Continuous Light Dark Problem with a few modifications. We extend it to a 2D domain and place a single ``light beacon'' in the continuous world. The agent's goal is to get to location $(0,0)$ and execute the terminal action - $Null$. Executing it within a small radius from $(0,0)$ will give the agent a reward of 200, and executing it outside the radius will yield a negative reward of -200. The agent can move in eight evenly spread directions $\mathcal{A} = \{  \rightarrow, \nearrow, \uparrow, \nwarrow,\leftarrow,  \swarrow,   \downarrow, \searrow,  Null\}$. The multi-objective reward function is $r(b,a,z,b') = -\expt{x\sim b'}{\nrm{2}{x}} - \lambda \hat{\mathcal{H}}(b,a,z,b')$. Motion, observation, and initial belief are $\mot{a}{x}{\cdot} = \mathcal{N}(x + a, \Sigma_T)$, $\observ{x}{z}=\mathcal{N}(x, \min\{1.0, \parallel x-x^b\parallel^2_2\} \cdot \Sigma_O)$, $b_0 = \mathcal{N}(x_0, \Sigma_0)$ respectively. $x^b$ is the 2D location of the beacon and all covariance matrices are diagonal (i.e. $\Sigma = I\cdot \sigma^2$). Implementation is built upon the JuliaPOMDP package collection \cite{egorov2017pomdps}. The code is attached alongside the supplementary manuscript. Extensive experiments confirm the advantage of our approach. We experiment with ten different configurations (rows of Table~\ref{tbl:run_times}) that differ in $m$  (number of particles), $d$ (simulation depth), and $\#$iter (number of simulation iterations per planning session). \begin{table}
\centering	
\caption{Runtimes of SITH-PFT versus PFT-DPW. The rows are different configurations of the number of belief particles $m$, maximal tree depth $d$, and the number of iterations per planning session. Reported values are averaged over 25 simulations 10 planning sessions each, and presented with the standard errors. In all simulations SITH-PFT and PFT-DPW declared \emph{identical} actions as optimal and exhibited \emph{identical} belief trees in terms of connectivity and visitation counts.\\}
\label{tbl:run_times}
	\begin{tabular}{ c c c } 
		\toprule 
		($m$, $d$, \verb|#|iter.) & Algorithm  & planning time [sec]   \\ 
		\midrule 
		\multirow{2}{*}{ (50, 30, 200)} & PFT-DPW & $ 3.54 \pm 0.4  $ \\ 
		& SITH-PFT &  $ 2.96 \pm 0.49  $ \\ 
		\hline 
		\multirow{2}{*}{ (50, 50, 500)} & PFT-DPW & $ 9.82 \pm 1.31  $ \\ 
		& SITH-PFT &  $ 8.1 \pm 1.33  $ \\ 
		\hline 
		\multirow{2}{*}{ (100, 30, 200)} & PFT-DPW & $ 13.42 \pm 1.49  $ \\ 
		& SITH-PFT &  $ 10.77 \pm 1.73  $ \\ 
		\hline 
		\multirow{2}{*}{ (100, 50, 500)} & PFT-DPW & $ 35.06 \pm 4.44  $ \\ 
		& SITH-PFT &  $ 26.7 \pm 4.37  $ \\ 
		\hline 
		\multirow{2}{*}{ (200, 30, 200)} & PFT-DPW & $ 55.89 \pm 5.41  $ \\ 
		& SITH-PFT &  $ 39.46 \pm 7.09  $ \\ 
		\hline 
		\multirow{2}{*}{ (200, 50, 500)} & PFT-DPW & $ 142.14 \pm 12.39  $ \\ 
		& SITH-PFT &  $ 100.09 \pm 14.67  $ \\ 
		\hline 
		\multirow{2}{*}{ (400, 30, 200)} & PFT-DPW & $ 211.86 \pm 24.18  $ \\ 
		& SITH-PFT &  $ 160.36 \pm 31.02  $ \\ 
		\hline 
		\multirow{2}{*}{ (400, 50, 500)} & PFT-DPW & $ 570.13 \pm 45.48  $ \\ 
		& SITH-PFT &  $ 414.65 \pm 53.37  $ \\ 
		\hline 
		\multirow{2}{*}{ (600, 30, 200)} & PFT-DPW & $ 503.78 \pm 31.61  $ \\ 
		& SITH-PFT &  $ 374.0 \pm 44.23  $ \\ 
		\hline 
		\multirow{2}{*}{ (600, 50, 500)} & PFT-DPW & $ 1204.78 \pm 119.16  $ \\ 
		& SITH-PFT &  $ 912.92 \pm 116.08  $ \\ 
		\bottomrule 
	\end{tabular} 
\end{table}Each scenario comprises $10$ planning sessions i.e.~the agent performs up to $10$ planning-action executing iterations. We repeat each of the experiments 25 times. In all different configurations, we obtained significant speedup while achieving the exact same solution compared to PFT. Results are summed up in Table~\ref{tbl:run_times}. An illustration can be found in Fig.~\ref{fig:cl2d}. Note that SITH-PFT \ref{fig:sith-pft_cl2d} yields identical to PFT solution \ref{fig:pft_cl2d} while IPFT demonstrates severely degraded behavior. We remind the purpose of our work is to speed up the PFT approach when coupled with information-theoretic reward. Hence, due to space constraints and since the two algorithms produce identical belief trees and action at the end of each planning session, there is no point reporting the algorithms \emph{identical} performances (apart from planning time).
For our simulations, we used an 8 cores Intel(R) Xeon(R) CPU E5-1620 v4 with 128 GB of RAM working at 3.50GHz.

\section{Conclusions}
\label{sec:concl}
We presented a novel method to accelerate information-theoretic reward planning. Our approach is applicable with any converging to the reward bounds.  We provide thorough proofs that our method is entirely equivalent to PFT-DPW, yielding the same solution and belief tree in each planning step. Our experiments demonstrate that the technique is paramount in terms of computation time compared to PFT-DPW. In the worst-case scenario, the computation time is approaching the baseline. The limitation of our algorithm is that it leans on converging bounds, which are not trivial to derive and specific for a particular reward function.  
In addition, it requires slightly more caching than the baseline.

	\section*{Acknowledgments}
	
	This research was  supported by the Israel Science Foundation (ISF) and by 
	a donation from the Zuckerman Fund to the Technion Center for Machine Learning and Intelligent Systems (MLIS).
	
	\bibliographystyle{plainnat}


\begin{thebibliography}{15}
\providecommand{\natexlab}[1]{#1}
\providecommand{\url}[1]{\texttt{#1}}
\expandafter\ifx\csname urlstyle\endcsname\relax
  \providecommand{\doi}[1]{doi: #1}\else
  \providecommand{\doi}{doi: \begingroup \urlstyle{rm}\Url}\fi

\bibitem[{Boers} et~al.(2010){Boers}, {Driessen}, {Bagchi}, and
  {Mandal}]{Boers10fusion}
Y.~{Boers}, H.~{Driessen}, A.~{Bagchi}, and P.~{Mandal}.
\newblock Particle filter based entropy.
\newblock In \emph{2010 13th International Conference on Information Fusion},
  pages 1--8, 2010.
\newblock \doi{10.1109/ICIF.2010.5712013}.

\bibitem[Dressel and Kochenderfer(2017)]{Dressel17icaps}
Louis Dressel and Mykel~J. Kochenderfer.
\newblock Efficient decision-theoretic target localization.
\newblock In Laura Barbulescu, Jeremy Frank, Mausam, and Stephen~F. Smith,
  editors, \emph{Proceedings of the Twenty-Seventh International Conference on
  Automated Planning and Scheduling, {ICAPS} 2017, Pittsburgh, Pennsylvania,
  USA, June 18-23, 2017}, pages 70--78. {AAAI} Press, 2017.
\newblock URL
  \url{https://aaai.org/ocs/index.php/ICAPS/ICAPS17/paper/view/15761}.

\bibitem[Egorov et~al.(2017)Egorov, Sunberg, Balaban, Wheeler, Gupta, and
  Kochenderfer]{egorov2017pomdps}
Maxim Egorov, Zachary~N. Sunberg, Edward Balaban, Tim~A. Wheeler, Jayesh~K.
  Gupta, and Mykel~J. Kochenderfer.
\newblock {POMDP}s.jl: A framework for sequential decision making under
  uncertainty.
\newblock \emph{Journal of Machine Learning Research}, 18\penalty0
  (26):\penalty0 1--5, 2017.
\newblock URL \url{http://jmlr.org/papers/v18/16-300.html}.

\bibitem[Elimelech and Indelman(2018)]{Elimelech18ijrr_submitted}
Khen Elimelech and Vadim Indelman.
\newblock Simplified decision making in the belief space using belief
  sparsification.
\newblock \emph{Intl. J. of Robotics Research}, 12 2018.
\newblock Conditionally accepted.

\bibitem[Fischer and Tas(2020)]{Fischer20icml}
Johannes Fischer and Omer~Sahin Tas.
\newblock Information particle filter tree: An online algorithm for pomdps with
  belief-based rewards on continuous domains.
\newblock In \emph{Intl. Conf. on Machine Learning (ICML)}, Vienna, Austria,
  2020.

\bibitem[Garg et~al.(2019)Garg, Hsu, and Lee]{Garg19rss}
Neha~P Garg, David Hsu, and Wee~Sun Lee.
\newblock Despot-$\alpha$: Online pomdp planning with large state and
  observation spaces.
\newblock In \emph{Robotics: Science and Systems (RSS)}, 2019.

\bibitem[Hoerger et~al.(2019)Hoerger, Kurniawati, and Elfes]{Hoerger19isrr}
Marcus Hoerger, Hanna Kurniawati, and Alberto Elfes.
\newblock Multilevel monte-carlo for solving pomdps online.
\newblock In \emph{Proc. International Symposium on Robotics Research (ISRR)},
  2019.

\bibitem[Kochenderfer et~al.(2022)Kochenderfer, Wheeler, and
  Wray]{Kochenderfer22book}
M.~Kochenderfer, T.~Wheeler, and K.~Wray.
\newblock \emph{Algorithms for Decision Making}.
\newblock MIT Press, 2022.

\bibitem[Kocsis and Szepesv{\'a}ri(2006)]{Kocsis06ecml}
Levente Kocsis and Csaba Szepesv{\'a}ri.
\newblock Bandit based monte-carlo planning.
\newblock In \emph{European conference on machine learning}, pages 282--293.
  Springer, 2006.

\bibitem[Kurniawati et~al.(2008)Kurniawati, Hsu, and Lee]{Kurniawati08rss}
H.~Kurniawati, D.~Hsu, and W.~S. Lee.
\newblock {SARSOP}: Efficient point-based {POMDP} planning by approximating
  optimally reachable belief spaces.
\newblock In \emph{Robotics: Science and Systems (RSS)}, volume 2008, 2008.

\bibitem[Silver and Veness(2010)]{Silver10nips}
David Silver and Joel Veness.
\newblock Monte-carlo planning in large pomdps.
\newblock In \emph{Advances in Neural Information Processing Systems (NIPS)},
  pages 2164--2172, 2010.

\bibitem[Sunberg and Kochenderfer(2018)]{Sunberg18icaps}
Zachary Sunberg and Mykel Kochenderfer.
\newblock Online algorithms for pomdps with continuous state, action, and
  observation spaces.
\newblock In \emph{Proceedings of the International Conference on Automated
  Planning and Scheduling}, volume~28, 2018.

\bibitem[Sztyglic and Indelman(2021)]{Sztyglic21arxiv}
Ori Sztyglic and Vadim Indelman.
\newblock Online pomdp planning via simplification.
\newblock \emph{arXiv preprint arXiv:2105.05296}, 2021.

\bibitem[Ye et~al.(2017)Ye, Somani, Hsu, and Lee]{Ye17jair}
Nan Ye, Adhiraj Somani, David Hsu, and Wee~Sun Lee.
\newblock Despot: Online pomdp planning with regularization.
\newblock \emph{JAIR}, 58:\penalty0 231--266, 2017.

\bibitem[Zhitnikov and Indelman(2021)]{Zhitnikov21arxiv}
Andrey Zhitnikov and Vadim Indelman.
\newblock Probabilistic loss and its online characterization for simplified
  decision making under uncertainty.
\newblock \emph{arXiv preprint arXiv:2105.05789}, 2021.

\end{thebibliography}
	
\section{Supplementary}
\subsection{Information theoretic bounds}\label{sec:supp_bounds}	
\label{sec:append}
In this paper we consider the differential entropy approximation by \cite{Boers10fusion}. The approximation is w.r.t. belief $b_{k+1}$ and assumes the form $\hat{\mathcal{H}}(b_{k}, a_k, z_{k+1}, b_{k+1})$ (see full expression in Sec.~\ref{sec:time_complexity}). Further, we consider bounds over this approximation developed by \cite{Sztyglic21arxiv} upholding \eqref{eq:bounds}.
Specifically, 
\begin{align}\label{eq:immediate_bounds}
	&u \bydef -\log \left[\sum_i \probd_Z(z_{k+1}\sep x_{k+1}^i) w_k^i\right]  +\sum_{i \in \neg A^s_{k+1}} w_{k+1}^i\cdot\log\left[ \text{const} \cdot \probd_Z(z_{k+1}\sep x_{k+1}^i)\right] \nonumber\\
	&+\sum_{i \in A^s_{k+1}} w_{k+1}^i\cdot\log\left[\probd_Z(z_{k+1}\sep x_{k+1}^i) \sum_j\probd_T(x_{k+1}^i\sep x_k^j, a_k)w_k^j\right]\nonumber \\
	& \ell \bydef -\log \left[\sum_i \probd_Z(z_{k+1}\sep x_{k+1}^i) w_k^i\right] +\sum_i w_{k+1}^i\cdot\log\left[ \probd_Z(z_{k+1}\sep x_{k+1}^i)\sum_{j \in A^s_k} \probd_T(x_{k+1}^i \sep x_k^j, a_k)w_k^j\right], 
\end{align}
where $\text{const} = \maxim{\max}{x'} \ \mot{a}{x}{x'}$.

\subsection{Proofs}\label{sec:supp_proofs}
\subsubsection{Assumptions}
For the following proofs (Secs.~\ref{sec:tree_consis} and  \ref{sec:sol_consis}) assume we are using a \emph{converging and finite-time resimplification strategy} that satisfies Assumptions~\ref{as:converg_resimpl},\ref{as:finite_resimpl}.
\subsubsection{Proof for Theorem \ref{thm:tree_consis}}\label{sec:tree_consis}
\begin{proof} \label{proof:thm_1}
	We provide proof by induction on the belief trees structure.\\ 
	\textbf{Base:} Consider an initial given belief node $b_0$. No actions were taken and no observations were made. Thus, both PFT tree and SITH-PFT trees contain a single identical belief node and the claim holds.\\
	\textbf{Induction hypothesis:} Assume we are given two identical trees with $n$ nodes, generated by PFT and a SITH-PFT. The trees uphold the terms of Definition~\ref{def:tree_consis}.\\
	\textbf{Induction step:} Assume by contradiction that in the next simulation (expanding the belief tree by one belief node by definition) different nodes were added to the trees. Thus, we got different trees.\\
	Two different scenarios are possible:

	\begin{case}\label{enum:case_1}
		The same action-observation sequence $a_0, z_1, a_1, z_2... a_m$ was chosen in both trees, but different nodes were added.
	\end{case}
	\begin{case}\label{enum:case_2}
		Different action-observation sequences were chosen for both trees and thus, we got different trees structure.
	\end{case}
	Case~\ref{enum:case_1} is not possible. Since the Induction hypothesis holds, the last action  $a_m$ was taken from the same node denoted $h'$ shared and identical to both trees. Next the same observation model is sampled for a new observation and a new belief node is added with a rollout emanating from it. The new belief nodes and the rollout are identical for both trees since both algorithms use the same randomization seed and the same observation and motion models.\\ \\
	Case~\ref{enum:case_2} must be true since we showed Case~\ref{enum:case_1} is false. There are two possible scenarios such that different action-observation sequences were chosen:
	\begin{subcases}
		\begin{case}\label{enum:case_2.1}
			At some point of the actions-observations sequence, different observations $z_i, z'_i$ were chosen.
		\end{case}
		\begin{case}\label{enum:case_2.2}
			At some point of the actions-observations sequence, PFT chose action $a^{\dagger}$ while SITH-PFT chose a different action, $\tilde{a}$, or even got stuck without picking any action.
		\end{case}

	Case~\ref{enum:case_2.1} is not possible since if new observations were made, they are the same one by reasons contradicting Case~\ref{enum:case_1} . If we draw existing observations (choose some observation branch down the tree) the same observations are drawn since they are drawn with the same random seed and from the same observations ``pool''. It is the same ``pool'' since the Induction hypothesis holds.\\ \\
	Case~\ref{enum:case_2.2} must be true since we showed Case~\ref{enum:case_2.1} is false, i.e., when both algorithms are at the identical node denoted as $h$ PFT chooses action $a^{\dagger}$, while SITH-PFT chooses a different action, $\tilde{a}$, or even got stuck without picking any action.
	Specifically, PFT chooses action $a^{\dagger} = \maxim{\argmax}{a}\ \text{UCB}$ and SITH-PFT's candidate action is $\tilde{a} = \maxim{\argmax }{a \in \mathcal{A}}  \ \lucb(ha)$.\\
	Three different scenarios are possible:

	\begin{subcases}
		\begin{case}\label{enum:case_2.2.1}
			the $\uucb, \lucb$ bounds over $h\tilde{a}$ were tight enough and $\tilde{a}$ was chosen such that $a^{\dagger} \neq \tilde{a}$.
		\end{case}
		\begin{case}\label{enum:case_2.2.2}
			SITH-PFT is stuck in an infinite loop. It can happen if the $\uucb, \lucb$ bounds over $h\tilde{a}$, and at least one of its sibling nodes $ha$, are not tight enough. However, all of the tree nodes are at the maximal simplification level. Hence, resimplification is triggered over and over without it changing anything.
		\end{case}
	\end{subcases}
\end{subcases}

	Case~\ref{enum:case_2.2.1} is not possible since the bounds are analytical (always true) and converge to the actual reward ($\lucb=\text{UCB}=\uucb$) for the maximal simplification level.\\ \\
	Case~\ref{enum:case_2.2.2} is not possible. If the bounds are not close enough to make a decision, resimplification is triggered. Each time some $ha$ node - sibling to $h\tilde{a}$ and maybe even $h\tilde{a}$ itself is chosen in \textit{Select Best} to over-go resimplification. According to Assumption.~\ref{as:converg_resimpl} and Assumption.~\ref{as:finite_resimpl}, after some finite number of iterations for all of the sibling $ha$ nodes (including $h\tilde{a}$) it holds $\lucb(ha)=\text{UCB}(ha)=\uucb(ha)$ and some action can be picked. If different actions have identical values we choose one by the same rule UCB picks actions with identical values (e.g. lower index/random).\\ \\
	Now, since Case~\ref{enum:case_2.2.2} is false, after some finite number of resimplification iterations, SITH-PFT will stop with bounds sufficient enough to make a decision. And since Case~\ref{enum:case_2.2.1} is false it holds that $a^{\dagger} = \tilde{a}$. Thus we get a contradiction and the proof is complete.
\end{proof}

\subsubsection{Proof for Theorem \ref{thm:sol_consis}:}\label{sec:sol_consis}
\begin{proof}
	Since the same tree is built according to Theorem~\ref{thm:tree_consis}, the only modification now is the final criteria at the end of the planning session at the root of the tree: $a^*=\maxim{\argmax}{a}\ Q(ha)$. Note we can set the exploration constant of UCB to $c=0$ and we get that UCB is just the $Q$ function. Thus if the bounds are not tight enough at the root to decide on an action, resimplification will be repeatedly called until SITH-PFT can make a decision. The action will be identical to the one chosen by UCB at PFT from similar arguments mentioned in the proof of Theorem~\ref{thm:tree_consis}, \ref{sec:tree_consis}. Note that additional final criteria for action selection could be introduced, but it would not matter since tree consistency is kept according to Theorem~\ref{thm:tree_consis} and the bounds converge to the actual immediate rewards and $Q$ estimations.
 \end{proof}

\subsubsection{Proof for Theorem \ref{thm:valid_strategy}}
We now prove the resimplification strategy described in section \ref{sec:chosen_startegy} is converging and finite-time resimplification strategy.

\begin{proof}[Proof: Converging resimplification strategy]
	Consider the condition for refinement of the bounds \eqref{eq:cond}. 
	Since $\frac{1}{d} g(ha)$ is the mean gap over all the nodes that are the descendants to $ha$, some of the nodes are above this mean gap, and some are under (accounting for the discount factor). We refine all the ones that are above. Further, for each descendant rollout, we refine one rollout node that is above the mean gap. If each time we refine all descendants belief nodes that are above the mean gap and one rollout node per descendant rollout (if it satisfies \eqref{eq:cond}), after one iteration the mean gap must decrease since there exists a node above the mean gap that got tighter. If there is no such node above the mean gap that means all the values $u' - \ell'$ are the same throughout the sub-tree and those values must be zero since the immediate bounds converge. Thus, the mean gap (and consequentially so does $\uucb(ha)-\lucb(ha)$) is getting smaller in each iteration unless it is already zero.
\end{proof}

\begin{proof}[Proof: Finite-time resimplification strategy]
	Similar to previous proof, in each iteration there exists a node above the mean gap that is chosen for refinement. There are no nodes above the gap only if throughout the sub-tree all the values $u' - \ell'$ are zero. This happens after a finite number of iterations since there is a finite number of nodes and a finite number of simplification levels. Since the bounds converge, at the maximal simplification level it holds $u'= \ell'\Rightarrow u'- \ell'=0$. Thus, after all nodes in the sub-tree got to the maximal simplification level it holds $\frac{1}{d} g(ha)=0$ and consequentially so does $\uucb(ha)-\lucb(ha)=0 \Rightarrow \uucb(ha)=\text{UCB}=\lucb(ha)$).
\end{proof}

\subsubsection{Time complexity analysis}\label{sec:time_complexity}
We turn to analyze the time complexity of our method using the chosen bounds \eqref{eq:immediate_bounds}.
We assume the significant bottleneck is querying the motion and observation models $\mot{a}{x}{x'}, \observ{x}{z}$ respectively.
Assume the belief is approximated by a set of $m$ weighted particles,
\begin{align}
	b = \{x^i, w^i\}_{i=1}^m. \label{eq:belief_particles}
\end{align}
Consider the \cite{Boers10fusion} differential entropy approximation for belief at time $k+1$,
\begin{align}
	&\hat{\mathcal{H}}(b_{k}, a_k, z_{k+1}, b_{k+1}) \bydef \underbrace{\log \left[\sum_i \observ{ x_{k+1}^i}{z_{k+1}} w_k^i\right]}_{a} + \label{eq:term_a} \\ 
	&\underbrace{\sum_i w_{k+1}^i\cdot\log\left[ \observ{ x_{k+1}^i}{z_{k+1}}\sum_{j} \mot{a_k}{x_k^j}{x_{k+1}^i}w_k^j\right]}_{b} \label{eq:term_b}
\end{align}
Denote the time complexity to query the observation and motion models a single time as $t_{obs}, t_{mot}$ respectively.
It is clear from \eqref{eq:belief_particles}, \eqref{eq:term_a} (term a) and, \eqref{eq:term_b} (term b) that:
\begin{align}
	\forall b \text{ as in \eqref{eq:belief_particles} }  \Theta(\hat{\mathcal{H}}(b)) = \Theta(m\cdot t_{obs} + m^2 \cdot t_{mot}).\label{eq:boers_complexity}
\end{align}
Since we share calculation between the bounds, the bounds' time complexity, for some level of simplification $s$, based  on \cite{Sztyglic21arxiv}, is:
\begin{align}
	\Theta(\ell^s + u^s) = \Theta(m\cdot t_{obs} + m^s\cdot m \cdot t_{mot}),\label{eq:bounds_complexity}
\end{align}
where $m^s$ is the size of the particles subset that is currently used for the bounds calculations, e.g.~$m^s= |A^s|$ ($A^s$ is as in \eqref{eq:immediate_bounds}) and $\ell^s, u^s$ denotes the immediate upper and lower bound using simplification level $s$. Further, we remind the simplification levels are discrete, finite, and satisfy
\begin{align}
	s \in \{1, 2, \ldots, M\},  \ \ \ell^{s=M} = \hat{\mathcal{H}}=u^{s=M}. \label{eq:simpl_levels}
\end{align}

Now, assume we wish to tighten $\ell^s, u^s$ and move from simplification level $s$ to $s+1$. 
Since the bounds are updated incrementally (as introduced by \cite{Sztyglic21arxiv}), when moving from simplification level $s$ to $s+1$ the only additional data we are missing are the new values of the observation and motion models for the newly added particles. Thus, we get that the time complexity of moving from one simplification level to another is:
\begin{align}
	\Theta(\ell^s + u^s \rightarrow \ell^{s+1} + u^{s+1}) = \Theta((m^{s+1}-m^s)\cdot m \cdot t_{mot}),\label{eq:increase_bounds_complexity}
\end{align}
where $	\Theta(\ell^s + u^s \rightarrow \ell^{s+1} + u^{s+1})$ denotes the time complexity of updating the bounds from one simplification level to the following one. Note the first term from \eqref{eq:bounds_complexity}, $m\cdot t_{obs}$, is not present in \eqref{eq:increase_bounds_complexity}. This term has nothing to do with simplification level $s$ and it is calculated linearly over all particles $m$.  Thus, it is calculated once at the beginning (initial/lowest simplification level).

We can now deduce using \eqref{eq:bounds_complexity} and \eqref{eq:increase_bounds_complexity}
\begin{align}
	\Theta(\ell^{s+1} + u^{s+1}) = \Theta(\ell^s + u^s) +  \Theta(\ell^s + u^s \rightarrow \ell^{s+1} + u^{s+1}).\label{eq:bounds_update_complexity}
\end{align}
Finally, using \eqref{eq:boers_complexity}, \eqref{eq:bounds_complexity}, \eqref{eq:simpl_levels}, \eqref{eq:increase_bounds_complexity}, and \eqref{eq:bounds_update_complexity}, we come to the conclusion that if at the end of a planning  session, a node's $b$ simplification level was $1\leq s \leq M$ than the time complexity saved for that node is 
\begin{align}
	\Theta((m-m^s)\cdot m \cdot t_{mot}) \label{eq:time_saved}.
\end{align}
This makes perfect sense since if we had to resimplify all the way to the maximal level we get $s=M\Rightarrow m^{s=M}=m$ and by substituting $m^s=m$ in \eqref{eq:time_saved} we saved no time at all.

To conclude, the total speedup of the algorithm is dependent on how many belief nodes' bounds were not resimplified to the maximal level. The more nodes we had at the end of a planning session with lower simplification levels, the more speedup we get according to \eqref{eq:time_saved}.

\subsubsection{Additional resimplification strategies}\label{sec:supp_more_strategies}
We note that the proofs for Theorems~\ref{thm:tree_consis}, \ref{thm:sol_consis} depends on our resimplification strategy \ref{sec:chosen_startegy}. That is, additional strategies can be introduced as long as they satisfy Assumptions \ref{as:converg_resimpl}, and \ref{as:finite_resimpl}. To clarify, a simple example of a converging and finite-time resimplification strategy would be to refine the bounds of all nodes (belief tree nodes and rollout nodes) that are descendants to the belief-action node $ha$ that was chosen for resimplification at \textit{Select Best} procedure. Naturally, there will always be a node that got tightened (unless all bounds are already equal); thus, Assumption.~\ref{as:converg_resimpl} is satisfied. Further, after a finite time, all nodes in the sub-tree got to the maximal level of simplification, and the bounds converged. Thus, Assumption.~\ref{as:finite_resimpl} is satisfied. Note that using this brute-force strategy can result in many unnecessary resimplifications. So, the potential speed-up may decrease but in the worst case, SITH-PFT will still yield the same time complexity as PFT.

\end{document}